\documentclass[10pt,twocolumn,letterpaper]{article}

\usepackage{cvpr}              

\usepackage{graphicx}
\usepackage{amsmath}
\usepackage{amssymb}
\usepackage{booktabs}
\usepackage{array}
\usepackage{arydshln}
\usepackage{multirow}
\usepackage{pifont}
\usepackage{tabu}
\usepackage[accsupp]{axessibility}

\usepackage[pagebackref,breaklinks,colorlinks]{hyperref}

\usepackage[capitalize]{cleveref}
\crefname{section}{Sec.}{Secs.}
\Crefname{section}{Section}{Sections}
\Crefname{table}{Table}{Tables}
\crefname{table}{Tab.}{Tabs.}

\def\confName{CVPR}
\def\confYear{2023}

\begin{document}

\title{GANHead: Towards Generative Animatable Neural Head Avatars}

\author{
        Sijing Wu$^{1}$
	~~
        Yichao Yan$^{2*}$
	~~ 
	Yunhao Li$^{1}$
	~~ 
	Yuhao Cheng$^{2}$ \\
        Wenhan Zhu$^{2}$
        ~~
        Ke Gao$^{3}$
        ~~    
        Xiaobo Li$^{3}$
        ~~
	Guangtao Zhai$^{1,2*}$ \\
        $^{1}$Institute of Image Communication and Network Engineering, Shanghai Jiao Tong University \\
        $^{2}$MoE Key Lab of Artificial Intelligence, AI Institute, Shanghai Jiao Tong University \\
        $^{3}$Alibaba Group \\
{\tt\small \{wusijing, yanyichao, lyhsjtu, chengyuhao, zhuwenhan823, zhaiguangtao\}@sjtu.edu.cn} \\
{\tt\small \{gaoke.gao, xiaobo.lixb\}@alibaba-inc.com}
}

\twocolumn[{
    \renewcommand\twocolumn[1][]{#1}%
    \maketitle
    \begin{center}
    \includegraphics[width=\textwidth]{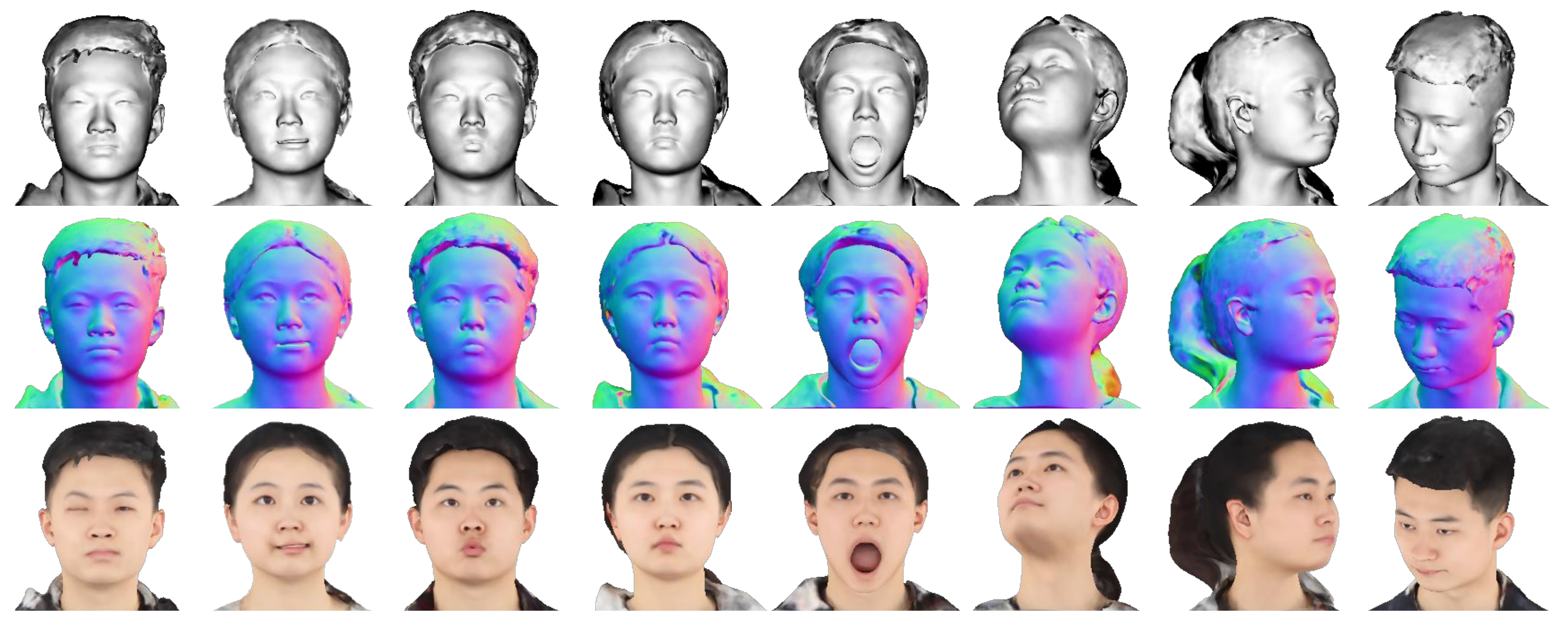}
    \captionof{figure}{We propose GANHead to generate diverse head avatars with complete geometry and realistic texture. The generated avatars can be deformed to target poses and expressions through FLAME \cite{li2017learning} parameters, which generalize well to unseen poses and expressions.}
    \label{fig:teaser}
    \end{center}
}]


\let\thefootnote\relax\footnotetext{$^*$Corresponding author}
\let\thefootnote\relax\footnotetext{Project page: \url{https://wsj-sjtu.github.io/GANHead/}}

\begin{abstract}
To bring digital avatars into people's lives, it is highly demanded to efficiently generate complete, realistic, and animatable head avatars. This task is challenging, and it is difficult for existing methods to satisfy all the requirements at once. 
To achieve these goals, we propose GANHead (\textbf{G}enerative \textbf{A}nimatable \textbf{N}eural \textbf{Head} Avatar), a novel generative head model that takes advantages of both the fine-grained control over the explicit expression parameters and the realistic rendering results of implicit representations. 
Specifically, GANHead represents coarse geometry, fine-gained details and texture via three networks in canonical space to obtain the ability to generate complete and realistic head avatars. To achieve flexible animation, we define the deformation filed by standard linear blend skinning (LBS), with the learned continuous pose and expression bases and LBS weights. This allows the avatars to be directly animated by FLAME \cite{li2017learning} parameters and generalize well to unseen poses and expressions. Compared to state-of-the-art (SOTA) methods, GANHead achieves superior performance on head avatar generation and raw scan fitting.
\end{abstract}

\section{Introduction}
\label{sec:intro}

How to efficiently generate photorealistic and animatable head avatars without manual effort is an open problem in computer vision and computer graphics, which has numerous applications in VR/AR, games, movies, and the metaverse. In these applications, it is desirable that the head avatar models fulfill the following requirements: (1) \textbf{Complete}, \ie, the 3D model can cover the entire head including the frontal face, the back of head, and the hair region; (2) \textbf{Realistic}, where the avatar is expected to display vivid texture and detailed geometry; (3) \textbf{Animatable}, \ie, the avatar is supposed to be fully riggable over poses and expressions, and can be controlled with low-dimensional parameters; (4) \textbf{Generative} model can be more flexibly applied to various downstream tasks, therefore, the head avatar model is preferred to be generative rather than discriminative for large-scale content generation tasks.

We investigate the research on neural head avatars and summarize previous works in Tab. \ref{tab:summary}.
3D morphable models (3DMMs) built from registered meshes have been widely employed to model head avatars. Principal component analysis (PCA) is applied to shape and texture, and novel subjects can be generated by sampling the coefficients of PCA bases. However, registering real-world raw scans to a template mesh with fixed topology is non-trivial, and it is difficult to define a fixed topology for complex regions like hair. As a result, most of these methods only model the facial region \cite{blanz1999morphable,cao2013facewarehouse,booth20163d,gerig2018morphable,egger2018occlusion,booth2018large,tran2019towards,liu20193d}, while a few cover the full head without hair \cite{li2017learning,ranjan2018generating,ploumpis2019combining,dai2020statistical,hifi3dface2021tencentailab}. Moreover, the oversimplification of PCA makes the models lack of realism.

In parallel with explicit meshes, implicit representations have been utilized to approximate complex surfaces. Some discriminative models \cite{ichim2015dynamic,gafni2021dynamic,park2021nerfies,zheng2022avatar,lombardi2021mixture} successfully model the complete head geometry with realistic texture. However, these methods can only be applied to the reconstruction task, incapable of generating new samples.
Meanwhile, 3D-aware GANs based on implicit representations \cite{chan2021pi,chan2022efficient,or2022stylesdf,zhao2022generative} can generate multi-view-consistent frontal face images. Nevertheless, the heads are still incomplete. In addition, it is difficult to animate the neural head avatars generated by 3D-aware GANs. 
Recently, several implicit generative models \cite{yenamandra2021i3dmm,zhuang2021mofanerf,hong2022headnerf,wang2022morf} achieve realistic and animatable head avatars. However, these models either cannot generate complete head with satisfactory geometry \cite{zhuang2021mofanerf,hong2022headnerf}, or can only be animated implicitly via the learned latent codes \cite{yenamandra2021i3dmm}, which is inconvenient and limits the generalization ability to unseen poses and expressions.

\begin{table}
\renewcommand\arraystretch{1.2}

\setlength{\belowcaptionskip}{-5pt}
  \footnotesize
  \setlength{\tabcolsep}{2pt}

\begin{tabular}{cccccc}
\toprule
Scheme & Methods   & Complete & Realistic & Animatable & Generative\\ \hline
\multirow{2}{*}{\shortstack{Explicit \\ 3DMMs}}    & \multirow{2}{*}{\cite{blanz1999morphable,gerig2018morphable,tran2019learning,hifi3dface2021tencentailab}}                          & \multirow{2}{*}{\ding{55}}        & \multirow{2}{*}{\ding{55}}         & \multirow{2}{*}{\ding{51}}      & \multirow{2}{*}{\ding{51}}  \\
\\
\hdashline
                     
\multirow{2}{*}{\shortstack{3D-aware \\ GANs}}    & \multirow{2}{*}{\cite{chan2021pi,chan2022efficient,or2022stylesdf,zhao2022generative}}                     & \multirow{2}{*}{\ding{55}}        & \multirow{2}{*}{\ding{51}}         & \multirow{2}{*}{\ding{55}}     & \multirow{2}{*}{\ding{51}}           \\
\\
\hdashline
\multirow{2}{*}{\shortstack{Personalized \\ Avatars}} & \cite{zheng2022avatar,grassal2022neural}                                                     & \ding{51}        & \ding{51}         & \ding{51}      & \ding{55}          \\
& \cite{wang2021learning,gafni2021dynamic,park2021nerfies}                                         & \ding{55}        & \ding{51}         &  $\triangle$     & \ding{55}     \\ \hdashline
\multirow{4}{*}{\shortstack{Implicit \\ Head Models}} 

& \cite{zhuang2021mofanerf}  & \ding{55}  & \ding{51}         & $\triangle$     & \ding{51}      \\
& \cite{hong2022headnerf}     & \ding{55}  & \ding{51}  &  \ding{51} & \ding{51}  \\
& \cite{yenamandra2021i3dmm} & \ding{51}  & \ding{51}  &  $\triangle$  & \ding{51}  \\ 
& Ours & \ding{51}  & \ding{51}  & \ding{51}  & \ding{51}   \\ \bottomrule
\end{tabular}
\caption{A summary of current head avatar methods. $\triangle$ denotes that the head avatar can only be animated implicitly via the learned latent codes, and cannot generalize well to unseen expressions. }
\label{tab:summary}
\end{table}

It is natural to ask a question: can we build a model that can generate diverse realistic head avatars, and meanwhile be compatible with the animation parameters of the common parametric face model (such as FLAME \cite{li2017learning})?
In this work, we propose a generative animatable neural head avatar model, namely, GANHead, that simultaneously fulfills these requirements.
Specifically, GANHead represents the 3D head implicitly with neural occupancy function learned by MLPs, where coarse geometry, fine-gained details and texture are respectively modeled via three networks. Supervised with unregistered ground truth 3D head scans, all these networks are defined in canonical space via auto-decoder structures that are conditioned by shape, detail and color latent codes, respectively. This framework allows GANHead to achieve \textbf{complete} and \textbf{realistic} generation results, while yielding desirable \textbf{generative} capacity.

The only remaining question is how to control the implicit representation with animation parameters?  
To answer this question, we extend the multi-subject forward skinning method designed for human bodies \cite{chen2022gdna} to human faces, enabling our framework to achieve flexible \textbf{animation} explicitly controlled by FLAME \cite{li2017learning} pose and expression parameters. Inspired by IMAvatar \cite{zheng2022avatar}, the deformation field in GANHead is defined by standard vertex based linear blend skinning (LBS) with the learned pose-dependent corrective bases, the linear blend skinning weights, and the learned expression bases to capture non-rigid deformations. In this way, GANHead can be learned from textured scans, and no registration or canonical shapes are needed.

Once GANHead is trained, we can sample shape, detail and color latent codes to generate diverse textured head avatars, which can then be animated flexibly by FLAME parameters with nice geometry consistency and pose/expression generalization capability.
We compare our method with the state-of-the-art (SOTA) complete head generative models, and demonstrate the superiority of our method.

In summary, our main contributions are:
\begin{itemize}
    \vspace{-0.5mm}
    \item We propose a generative animatable head model that can generate complete head avatars with realistic texture and detailed geometry.
    \vspace{-0.5mm}
    \item The generated avatars can be directly animated by FLAME \cite{li2017learning} parameters, robust to unseen poses and expressions.
    \vspace{-0.5mm}
    \item The proposed model achieves promising results in head avatar generation and raw scan fitting compared with SOTA methods.
\end{itemize}

\section{Related Work}
\label{sec:related work} 

\noindent\textbf{Explicit Face and Head Morphable Models.} Explicit representation is wildly used for 3D face modeling, which is built by performing Principal Component Analysis (PCA) on numerous registered 3D facial scans and represents a 3D face as the linear combination of a set of orthogonal bases.
Blanz and Vetter \cite{blanz1999morphable} first proposed the concept of 3D Morphable Face Model (3DMM). Since then, many efforts \cite{cao2013facewarehouse,abrevaya2018multilinear,booth20163d,gerig2018morphable,booth2018large} have been devoted to improve the performance of 3DMM by either improving the quality of captured face scans or the structure of 3D face model. 
Considering the limited representation power of traditional 3D Morphable Models and the difficulty of acquiring registered 3D data, deep learning based 3DMMs appeared \cite{tran2019learning,tran2019towards,tewari2019fml,tewari2021learning}, which learn 3D priors from 2D face images or videos with the help of differentiable rendering.
However, these methods \cite{blanz1999morphable,paysan20093d,cao2013facewarehouse,booth20163d,gerig2018morphable,booth2018large} can only model the facial region.

Recently, some 3D Morphable Models that can represent the entire head have been proposed \cite{li2017learning,ranjan2018generating,ploumpis2019combining,dai20173d, dai2020statistical,hifi3dface2021tencentailab}. For example, Li \emph{et al.} \cite{li2017learning} propose the FLAME model which represents 3D head by rotatable joints and linear blend skinning. 
Although these methods can model the entire head, they still cannot model the hair region since it is hard to define a fixed topology for complex regions like hair and register the raw scan to it, while our model has the ability to generate complete head avatars with diverse hairstyles.

\vspace{2mm}
\noindent\textbf{Implicit Face and Head Models.} In parallel with explicit meshes, implicit representations \cite{mildenhall2021nerf,park2019deepsdf,mescheder2019occupancy,oechsle2021unisurf} can also be used to model 3D shapes \cite{park2019deepsdf,mescheder2019occupancy}. Park \emph{et al.} \cite{park2019deepsdf} propose DeepSDF to represent shape using signed distance function predicted by an autodecoder. Since then, implicit representations have become popular in 3D modeling, as well as 3D face and head modeling \cite{ramon2021h3d,xu2022point}, since implicit representations are better at modeling complex surfaces and realistic textures. 
Many works \cite{ramon2021h3d,kellnhofer2021neural,wang2021prior} successfully reconstruct high fidelity static heads which cannot be animated.
Recent works \cite{gafni2021dynamic,park2021nerfies,zheng2022avatar, raj2021pixel} recover animatable realistic head avatars from monocular RGB videos, but needed to train a model for each person.
In addition, 3D-aware GANs \cite{chan2021pi,chan2022efficient,or2022stylesdf,zhao2022generative,gu2021stylenerf} are proposed to generate multi-view-consistent static frontal face images, but failed to extract complete head meshes (including the back of the head) due to the lack of 3D supervision.

Recently, several implicit generative models \cite{yenamandra2021i3dmm,zhuang2021mofanerf,hong2022headnerf} are proposed to achieve realistic and animatable head avatars. Hong \emph{et al.} \cite{hong2022headnerf} propose the first NeRF-based parametric human head model which controls the rendering pose, identity and expression by corresponding latent codes. Yenamandra \emph{et al.} \cite{yenamandra2021i3dmm} propose i3DMM, a deep implicit 3D morphable model containing entire heads and can be animated by learned latent codes. However, these models either can not generate complete head meshes with satisfactory geometry \cite{zhuang2021mofanerf,hong2022headnerf}, or can only be animated implicitly via the learned latent codes \cite{yenamandra2021i3dmm}, which limits the generalization ability to unseen poses and expressions. In contrast, our method can generate animatable head avatars with complete geometry and realistic texture using implicit representation, which can also be generalized well to unseen poses and expressions.

\begin{figure*}[t]
\centering
\includegraphics[scale=0.55]{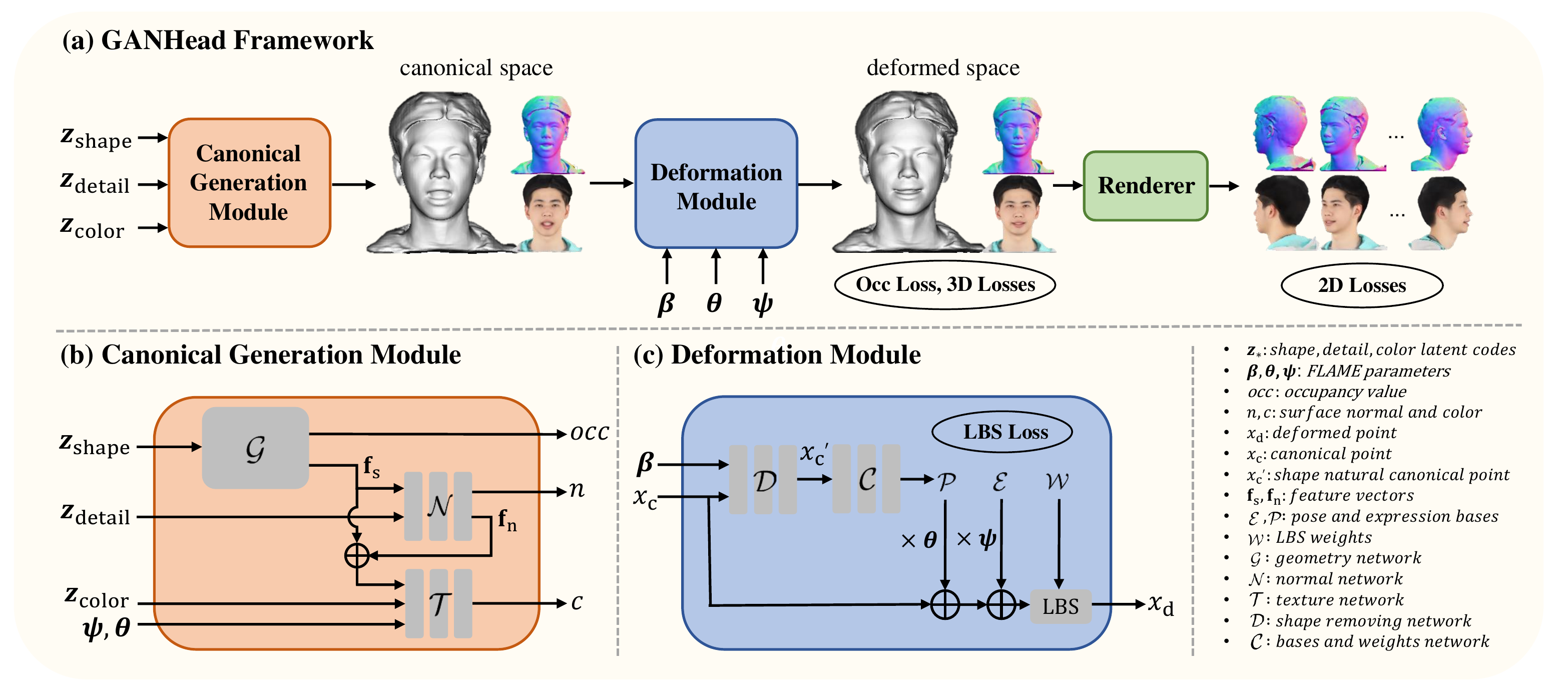}
\caption{\textbf{Method overview.} Given shape, detail and color latent codes, the canonical generation model outputs coarse geometry and detailed normal and texture in canonical space. The generated canonical head avatar can then be deformed to target pose and expression via the deformation module. In the first training stage, occupancy values of the deformed shapes are used to calculate the occupancy loss, along with the LBS loss, to supervise the geometry network and the deformation module. In the second stage, the deformed textured avatars are rendered to 2D RGB images and normal maps, together with the 3D color and normal losses, to supervise the normal and texture networks.}

\vspace{-2mm}
\label{fig:2}
\end{figure*}

\section{Method}

In this work, we propose GANHead, a generative model learned from unregistered textured scans. Once GANHead is trained, complete and realistic head avatars that are ready for animation can be obtained by sampling three latent codes. An overview of GANHead is illustrated in Fig.~\ref{fig:2}.

In this section, we first recap the deformation formulation of parametric head model FLAME \cite{li2017learning}, and illustrate its important role in helping GANHead build a deformation module with good generalization ability to unseen poses and expressions (Section \ref{subsec:method_part1}).
Second, we introduce the canonical generation module (Section \ref{subsec:method_part2}) that generates diverse vivid head avatars in canonical space, followed by the deformation module (Section \ref{subsec:method_part3}) which deforms the generated avatars to new poses and expressions controlled by FLAME parameters. Finally, to train GANHead model from raw scans, the data pre-processing procedures, training strategy and losses are introduced in Section \ref{subsec:method_part4}.

\subsection{Preliminary: GANHead vs FLAME}
\label{subsec:method_part1}
FLAME \cite{li2017learning} is a wildly used parametric model that covers the entire head (without hair), which is deformed by:
\begin{equation}
\label{eq:flame1}
M(\boldsymbol{\beta},\boldsymbol{\theta},\boldsymbol{\psi})
= LBS(T_P(\boldsymbol{\beta},\boldsymbol{\theta},\boldsymbol{\psi}),J(\boldsymbol{\beta}),\boldsymbol{\theta},\mathcal{W}),
\end{equation}
where $\boldsymbol{\beta}$, $\boldsymbol{\theta}$ and $\boldsymbol{\psi}$ denote the shape, pose and expression parameters respectively. $LBS(\cdot)$ is the standard vertex based linear blend skinning (LBS) and $\mathcal{W}$ denotes the skinning weights. $J(\cdot)$ calculates joints location from mesh vertices, and $T_P$ is calculated by:
\begin{equation}
\label{eq:flame2}
T_P(\boldsymbol{\beta},\boldsymbol{\theta},\boldsymbol{\psi})
=\overline{T} + B_S(\boldsymbol{\beta};\mathcal{S}) + B_P(\boldsymbol{\theta};\mathcal{P}) + B_E(\boldsymbol{\psi};\mathcal{E}),
\end{equation}
where, $\overline{T}$ is the template head. $B_S(\boldsymbol{\beta};\mathcal{S})$, $B_P(\boldsymbol{\theta};\mathcal{P})$ and $B_E(\boldsymbol{\psi};\mathcal{E})$ are 
per-vertex offsets calculated by shape parameters $\boldsymbol{\beta}$, pose parameters $\boldsymbol{\theta}$ and expression parameters $\boldsymbol{\psi}$ with corresponding bases $\mathcal{S}$, $\mathcal{P}$ and $\mathcal{E}$.

Different from FLAME, our framework aims to model the complete head geometry (including hair) and realistic texture. Therefore, we employ implicit representation due to its flexibility, rather than the polygon mesh used in FLAME. Specifically, the textured canonical shape (\textit{i.e.} head with identity information in natural pose and expression) is represented by neural occupancy function learned by MLPs, which is controlled by the learned latent codes.

Although implicit representations are more powerful, it is difficult for them to deform and generalize to unseen poses and expressions. To address this, we combine the implicit representations with the fine-grained control modeled in FLAME \cite{li2017learning} to enjoy the merits of both sides. However, the number of vertices is not fixed in implicit representation, such that the original bases and LBS weights in the FLAME model cannot be directly used in our framework. To further tackle this issue, we utilize an MLP to learn continuous pose and expression bases, as well as the LBS weights. In order to control the avatars generated by GANHead using the same pose and expression parameters as the FLAME model, we calculate the ground truth by finding the nearest neighbors of the query points on the fitted FLAME surface, to supervise the learning of neural bases and weights.

\subsection{Canonical Generation Module}
\label{subsec:method_part2}

GANHead models head shape and texture in canonical space via the canonical generation module, and we further design a deformation module in Section \ref{subsec:method_part3} to make it controllable by pose $\boldsymbol{\theta}$ and expression $\boldsymbol{\psi}$ parameters which are consistent with FLAME \cite{li2017learning}. It is remarkable that canonical heads are defined as: heads with identity information in natural pose and expression.

The canonical generation module consists of three neural networks that represent the shape (including coarse shape and fine-grained normal) and texture respectively.

\noindent\textbf{Shape:} We model the canonical head shape as the 0.5 level set of the occupancy function predicted by the geometry network $\mathcal{G}$:
\begin{equation}
\label{eq:2}
\mathcal{G}(x_c,\boldsymbol{z}_{\rm{shape}}): \mathbb{R}^3 \times \mathbb{R}^{n_s} \to occ,\boldsymbol{\rm{f}}_s,
\end{equation}
where $x_c$ denotes the point in canonical space, and $\boldsymbol{z}_{\rm{shape}} \in \mathbb{R}^{n_s}$ is the shape latent code that conditions $\mathcal{G}$ to generate diverse shapes. ${\boldsymbol{\rm{f}}_s} \in \mathbb{R}^{n_{\rm{f}}}$ is a feature vector carrying shape information which is then used to help predict fine-grained surface normal and texture. 
$\mathcal{G}$ consists of a 3D style based feature generator followed by an MLP conditioned by the generated feature, similar to \cite{chen2022gdna}.

To model the details of the head, we use an MLP to predict the surface normal:
\begin{equation}
\label{eq:2}
 \mathcal{N}(x_c,\boldsymbol{z}_{\rm{detail}},\boldsymbol{\rm{f}}_s): \mathbb{R}^3 \times \mathbb{R}^{n_d} \times \mathbb{R}^{n_{\rm{f}}} \to n,\boldsymbol{\rm{f}}_n,
\end{equation}
where $\boldsymbol{z}_{\rm{detail}} \in \mathbb{R}^{n_d}$ is the detail latent code that controls the detail generation. $n$ is the predicted normal of the query point $x_c$, together with a feature vector $\boldsymbol{\rm{f}}_n \in \mathbb{R}^{n_{\rm{f}}}$ used for texture prediction. 

\noindent\textbf{Texture:} We model the head texture in canonical space via a texture MLP $\mathcal{T}$:
\begin{equation}
\label{eq:2}
 \mathcal{T}(x_c,\boldsymbol{z}_{\rm{color}},\boldsymbol{\rm{f}},\boldsymbol{\theta},\boldsymbol{\psi}): \mathbb{R}^3 \times \mathbb{R}^{n_c} \times \mathbb{R}^{2n_{\rm{f}}} \times \mathbb{R}^{15} \times \mathbb{R}^{50} \to c,
\end{equation}
where $\boldsymbol{z}_{\rm{color}} \in \mathbb{R}^{n_c}$ is the color latent code that enables controllable texture generation. $\boldsymbol{\rm{f}}=\boldsymbol{\rm{f}}_s \oplus \boldsymbol{\rm{f}}_n$ is the concatenation of the shape and normal feature vectors. $\boldsymbol{\theta}$ and $\boldsymbol{\psi}$ denote the pose and expression parameters, respectively, which are consistent with FLAME \cite{li2017learning}.

\subsection{Deformation Module}
\label{subsec:method_part3}
To achieve flexible deformation with 3D geometry consistency and good generalization to unseen poses and expressions, we design our deformation module upon the FLAME \cite{li2017learning} deformation field, as discussed in Section \ref{subsec:method_part1}.
The deformation module first predicts the continuous pose and expression bases, as well as LBS weights of the canonical points $x_c$, and then deforms them to $x_d$ via added per-vertex offsets followed by linear blend skinning (LBS).

The continuous bases and weights are predicted via an MLP:
\begin{equation}
\label{eq:cnet}
\mathcal{C}(x_c): \mathbb{R}^3 \to 
\mathcal{E},\mathcal{P},\mathcal{W},
\end{equation}
where $\mathcal{E} \in \mathbb{R}^{3 \times 50}$, $\mathcal{P} \in \mathbb{R}^{36 \times 3}$ and $\mathcal{W} \in \mathbb{R}^{5}$ are the predicted expression bases, pose-dependent corrective bases and LBS weights of each canonical point $x_c$. Different from \cite{zheng2022avatar}, we expect the network $\mathcal{C}$ can be used to multiple individuals like traditional parametric face models, rather than a single person. To this end, we define the network $\mathcal{C}$ in shape natural space by adding a shape removing network $\mathcal{D}$ in front of $\mathcal{C}$:
\begin{equation}
\label{eq:2}
\mathcal{D}(x_c,\boldsymbol{\beta}): \mathbb{R}^3 \times \mathbb{R}^{100} \to {x_c}',
\end{equation}
with ${x_c}'$ denotes the shape natural canonical point, and $\boldsymbol{\beta}$ is the shape parameter consistent with FLAME \cite{li2017learning}. Accordingly, Eq. \ref{eq:cnet} can be rewritten as:
\begin{equation}
\mathcal{C}({x_c}'): \mathbb{R}^3 \to 
\mathcal{E},\mathcal{P},\mathcal{W}.
\end{equation}

Once the continuous bases and weights are predicted, the canonical head avatar can be deformed to target pose and expression by adding offsets followed by performing standard linear blend skinning (LBS):
\begin{align}
\label{eq:deform}
&X_P(\boldsymbol{\theta},\boldsymbol{\psi})
=X_c + B_P(\boldsymbol{\theta};\mathcal{P}) + B_E(\boldsymbol{\psi};\mathcal{E}),  \\ \nonumber
&X_d(\boldsymbol{\beta},\boldsymbol{\theta},\boldsymbol{\psi}) = LBS(X_P(\boldsymbol{\theta},\boldsymbol{\psi}),J(\boldsymbol{\beta}),\boldsymbol{\theta},\mathcal{W}),
\end{align}
where $X_c=\{x_{c1}, \ldots ,x_{cn}\}$ and $X_d=\{x_{d1}, \ldots ,x_{dn}\}$ denote the sets of the canonical and deformed points respectively. $\boldsymbol{\beta}$, $\boldsymbol{\theta}$ and $\boldsymbol{\psi}$ are the shape, pose and expression parameters consistent with FLAME \cite{li2017learning}, which makes the generated avatars easy to animate by FLAME parameters. To be clear that $\mathcal{E}$, $\mathcal{P}$ and $\mathcal{W}$ are the predicted continuous bases and weights rather than the corresponding FLAME components in Eq. \ref{eq:flame1}.

As mentioned before, the shape, normal and texture networks are all defined in canonical space to learn more details and generalize well to unseen poses and expressions, which means that if we input canonical query points to the canonical generation module, the output is an head avatar in canonical space, while if we input the canonical correspondence of the deformed query points, we will obtain the occupancy values, normals and colors of the deformed head avatar. The canonical correspondence of the deformed points $X_d$ is obtained by iteratively finding the root of Eq. \ref{eq:deform} given deformed points $X_d$ \cite{chen2021snarf}.

\begin{figure*}[h]
\centering
\includegraphics[scale=0.77]{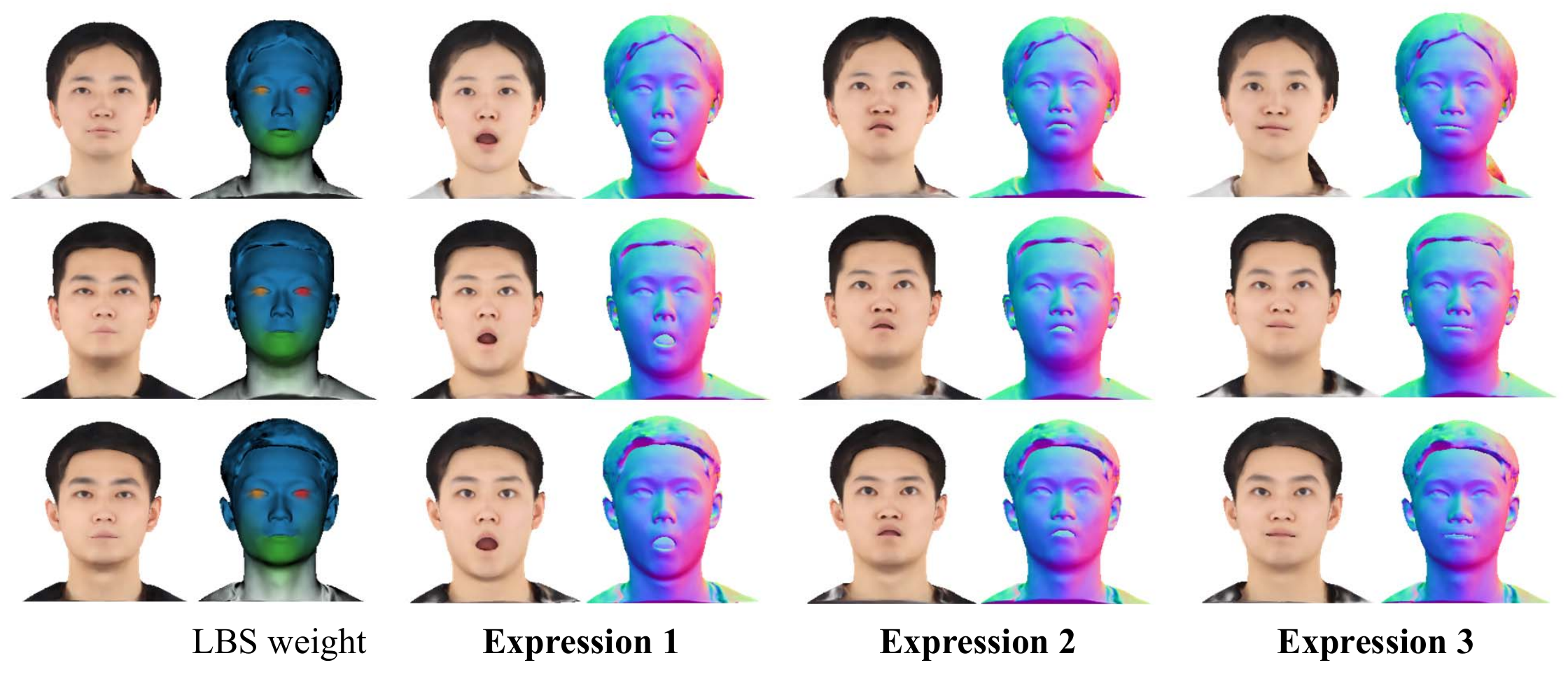}
\vspace{-4mm}
\caption{\textbf{Head avatars generation and animation.} 
We generate three head avatars by randomly sampling GANHead, and then deform them to the target expressions controlled by FLAME parameters. We show three samples, and they are all deformed to the same expressions. We also show the learned LBS weights of the canonical heads. Shapes are visualized as normal maps to highlight the geometry details.}

\vspace{-3mm}
\label{fig:generate_deform}
\end{figure*}

\subsection{Training}
\label{subsec:method_part4}
\noindent\textbf{Data:} We use the textured scans in FaceVerse-Dataset \cite{wang2022faceverse} to train our generative model GANHead. To obtain FLAME fitting results (shape parameters $\boldsymbol{\beta}$, pose parameters $\boldsymbol{\theta}$ and expression parameters $\boldsymbol{\psi}$) of the dataset for training, 3D facial landmarks are required for rigid alignment (\textit{i.e.} calculate the scale, translation and rotation factors to align the FLAME model with raw scans). To this end, we first calculate the 3D to 2D correspondence by rendering the scans to RGB images and depth images, then use Dlib \cite{king2009dlib} to detect the 2D landmarks and project them onto the 3D scans.

\noindent\textbf{Training strategy:} We train the GANHead model in two stages similar to gDNA \cite{chen2022gdna}: the coarse geometry network and deformation module are trained in the first stage, while the detail normal and texture networks are trained in the second stage. The canonical space is defined as the avatar opening its mouth slightly to help the model learn more details of the inner mouth.

\noindent\textbf{Losses:} In the first stage, We define the loss function as:
\begin{equation}
\label{eq:1}
\mathcal{L}_{stage1} = \mathcal{L}_{occ} + \lambda_{d}\mathcal{L}_{deshape} + \lambda_{l}\mathcal{L}_{lbs} + \lambda_{r}\mathcal{L}_{reg}.
\end{equation}
Specifically, $\mathcal{L}_{occ}$ measures the binary cross entropy between the predicted occupancy $\mathcal{G}(x_d,\boldsymbol{z}_{shape},\boldsymbol{\beta},\boldsymbol{\theta},\boldsymbol{\psi})$ and the ground truth occupancy $o_{gt}(x_d)$ of the sampled points $x_d$. $L_{deshape}$ supervises the shape removing network by enforcing the shape removing ability on FLAME vertices similar to \cite{chen2022gdna}:
\begin{equation}
\label{eq:1}
\mathcal{L}_{deshape} = \left \| \mathcal{D}(M(\boldsymbol{\beta},\boldsymbol{\theta}_0,\boldsymbol{\psi}_0),\boldsymbol{\beta}) - M(\boldsymbol{\beta}_0,\boldsymbol{\theta}_0,\boldsymbol{\psi}_0)\right \|_2^2,
\end{equation}
with $M(\cdot)$ and $\mathcal{D}(\cdot)$ denotes the FLAME model and shape removing network, respectively. $\boldsymbol{\beta}$, $\boldsymbol{\theta}$ and $\boldsymbol{\psi}$ are FLAME parameters, and the subscript $0$ indicates that the parameter is all zeros. 
The LBS loss $\mathcal{L}_{lbs}$ provides weak supervision for the learned LBS weights $\mathcal{W}$, pose bases $\mathcal{P}$ and expression bases $\mathcal{E}$ by constraining them to the correspondent FLAME components similar to \cite{zheng2022avatar}:
\begin{equation}
\label{eq:1}
\mathcal{L}_{lbs} = \left \| \mathcal{W} - \mathcal{W}_{gt} \right \|_2^2 +
\lambda_{p} \left \| \mathcal{P} - \mathcal{P}_{gt} \right \|_2^2 + 
\lambda_{e} \left \| \mathcal{E} - \mathcal{E}_{gt} \right \|_2^2,
\end{equation}
where $\mathcal{W}_{gt}$, $\mathcal{P}_{gt}$ and $\mathcal{E}_{gt}$ are the LBS weights, pose-dependent corrective bases and expression bases of FLAME model. It is remarkable that all the FLAME components are sampled to the same dimension as the query points $X_c$ by finding the nearest neighbours of the query points on the fitted
FLAME surface. Following \cite{chen2021snarf,chen2022gdna}, we add two auxiliary losses during the first training epoch (see Sup. Mat. for details). In addition, we employ a regularization term for the shape code via $\mathcal{L}_{reg}=\left \| \boldsymbol{z}_{\rm{shape}} \right \|_2^2$.

In the second stage, the training loss is defined as:
\begin{equation}
\label{eq:1}
\mathcal{L}_{stage2} = \lambda_{c}\mathcal{L}_{color} + \lambda_{n}\mathcal{L}_{normal} + \lambda_{r}\mathcal{L}_{reg}.
\end{equation}
The color loss $\mathcal{L}_{color}$ includes the 2D and 3D supervision of the texture:
\begin{equation}
\label{eq:1}
\mathcal{L}_{color} = \left \| I_r - I_{gt} \right \|_2^2 +
\lambda \left \| c - c_{gt} \right \|_2^2,
\end{equation}
where $I_{gt}$ and $I_r$ are the rendered RGB images of the ground truth scan and the output of GANHead respectively. $c_{gt}$ and $c$ denotes the ground truth color and the predicted color of the query points.
The normal loss also includes 2D and 3D supervision:
\begin{equation}
\label{eq:1}
\mathcal{L}_{normal} = \left \| N_r - N_{gt} \right \|_2^2 +
\lambda (1 - n_{gt}^{T} \cdot n),
\end{equation}
where $N_{gt}$ and $N_r$ are the rendered normal maps. $n_{gt}$ and $n$ are the normalized normal values of the query points, and we enforce them point to the same direction. Furthermore, we regularize the detail and color code via $\mathcal{L}_{reg}=\left \| \boldsymbol{z}_{\rm{detail}} \right \|_2^2 + \lambda \left \| \boldsymbol{z}_{\rm{color}} \right \|_2^2$.

\section{Experiments}
GANHead is proposed to generate diverse realistic head avatars that can be directly animated by FLAME \cite{li2017learning} parameters. In this section, we evaluate the superiority of GANHead in terms of the head avatar generation quality and the animation flexibility of the generated avatars. Furthermore, we also fit GANHead to unseen scans and compare the performance to SOTA animatable head models to evaluate its expressiveness.

\begin{figure*}[h]
\centering
\includegraphics[scale=0.53]{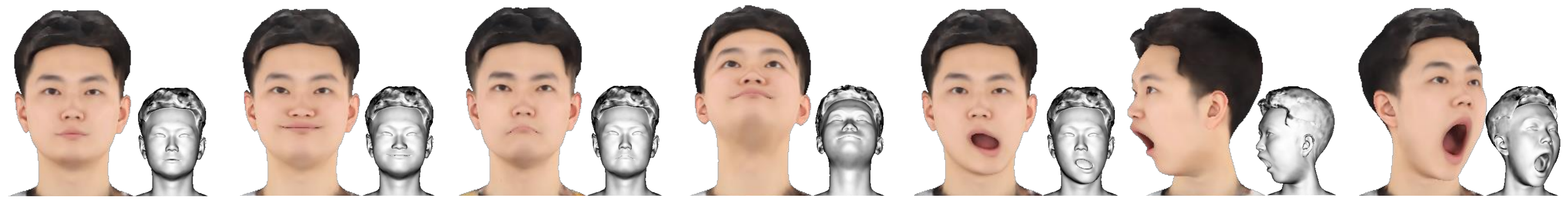}
\vspace{-3mm}
\caption{\textbf{Extreme poses and expressions.} We generate a head avatar and then deform it to several extreme poses and expressions which have never showed up in the training set.}

\vspace{-3mm}
\label{fig:extreme}
\end{figure*}

\subsection{Implementation Details} 
\textbf{Dataset:} We train our model on 2289 textured scans out of 2310 (110 identities, each with 21 expressions) from the training set of the FaceVerse-Dataset \cite{wang2022faceverse}. Scans of a subject with hat are removed to avoid interfering with the learning of hair. And the test set of the FaceVerse-Dataset (375 scans from 18 subjects) are used to evaluate the raw scan fitting. We further conduct experiments on a subset of Multiface dataset \cite{wuu2022multiface} to verify the generalization ability of our model on different datasets (See Sup. Mat.).

\textbf{Training details:} We use PyTorch to implement our model, and Adam optimiser is used for training. We train 250 epochs with a batch size of 32 for the first stage, and 200 epochs with a batch size of 4 for the second stage. The 3D and 2D correspondence are precomputed before the second stage. The whole training takes about 3 days on 4 NVIDIA 3090 GPUs. Please refer to Sup. Mat. for more details.

\begin{figure}[h]
\centering
\includegraphics[scale=0.445]{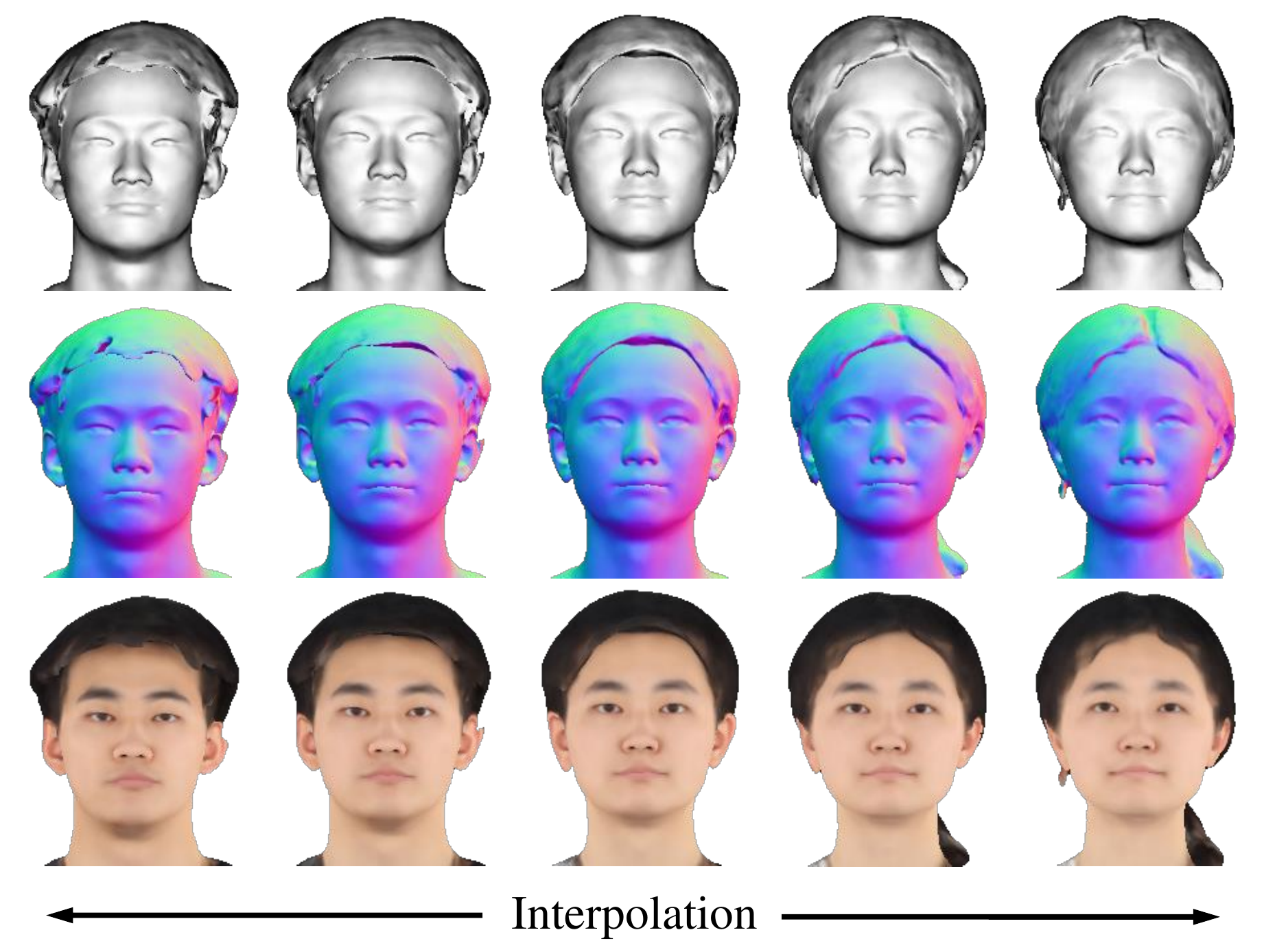}
\vspace{-3mm}
\caption{\textbf{Interpolation.} We interpolate the shape, detail and color latent codes between two samples, and show the geometry, detail and texture of the interpolation results.}
\vspace{-3mm}
\label{z_inter}
\end{figure}

\subsection{Generation and Animation Capacities}
\noindent\textbf{Random Generate:} We randomly sample the shape, detail and color latent codes to generate head avatars. The generated canonical head avatars and the visualization of their respective LBS weights are shown in Fig. \ref{fig:generate_deform} (the first column). We find that GANHead can generate diverse head avatars with detailed geometry and nice  textures.

\noindent\textbf{Deform to Target Poses and Expressions:} The avatars generated by GANHead can be easily animated controlled by FLAME \cite{li2017learning} parameters. Here we deform the generated avatars (The first column in Fig.~\ref{fig:generate_deform}) to the target expressions represented by FLAME \cite{li2017learning} parameters. The results show that the generated avatars can be well controlled by FLAME parameters, and the poses and expressions are well disentangled from the geometry.

\noindent\textbf{Deform to Unseen Extreme Expressions and Poses:} In GANHead, the deformation module controlled by FLAME~\cite{li2017learning} parameters makes the generated avatars generalize well to unseen poses and expressions. Here we generate a head avatar, and deform it to six extreme poses and expressions by changing the FLAME parameters $\boldsymbol{\theta}$ and $\boldsymbol{\psi}$. As shown in Fig. \ref{fig:extreme}, the generated avatar can be deformed to extreme expressions and poses that are not included in the training set, while displaying great geometry consistency. This is hard to achieve by previous implicit model.

\noindent\textbf{Latent Code Interpolation}: We interpolate the shape, detail and color latent codes of two samples that look vary different, as shown in Fig. \ref{z_inter}. We can see a smooth transition between the two samples.

\subsection{Ablation Study}
To validate the importance of each components of GANHead, we conduct ablation experiments on a subset (420 scans from 20 identities) of the training set. 

\begin{figure}[h]
\centering
\includegraphics[scale=0.35]{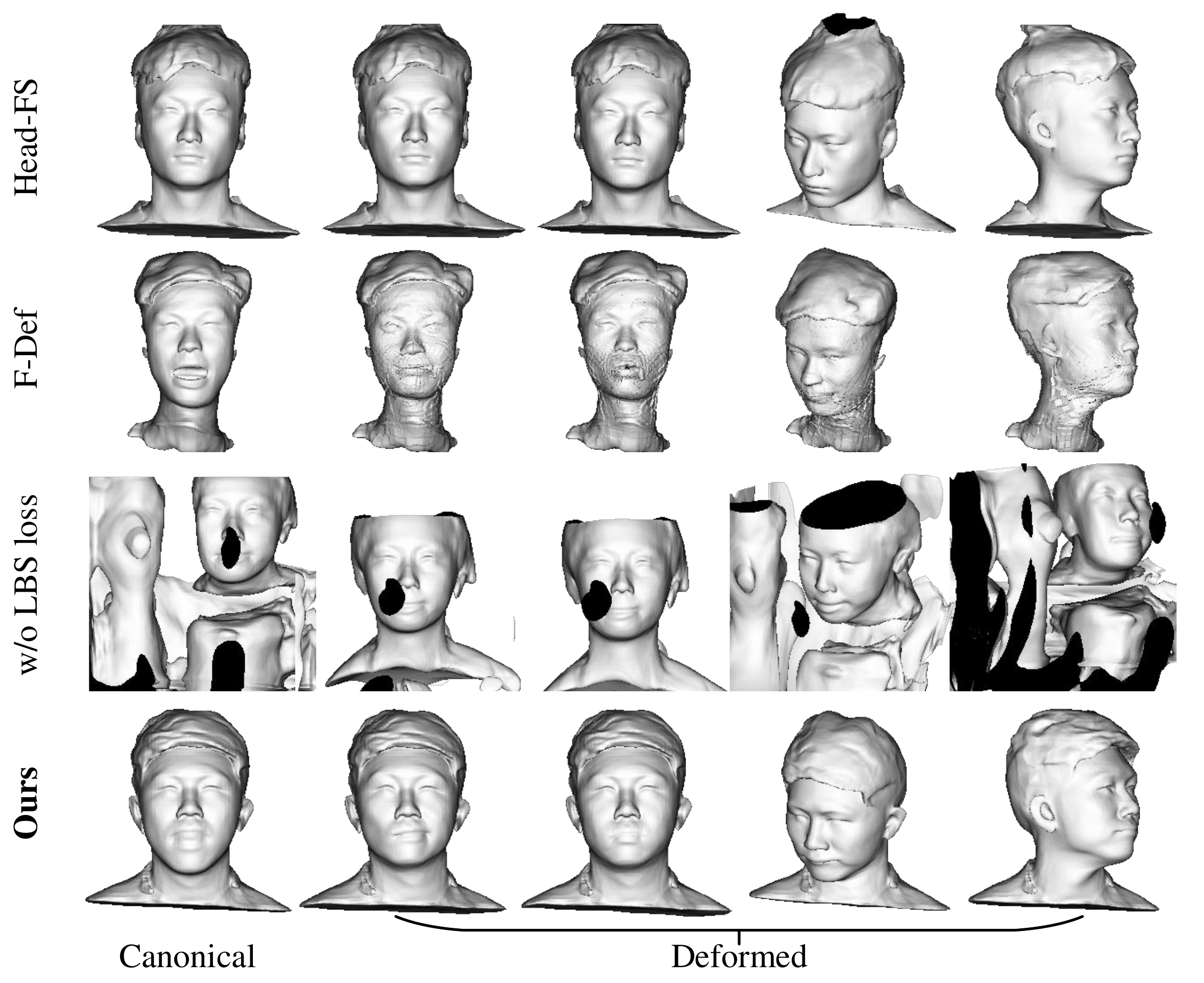}
\caption{\textbf{Comparison with baseline methods.} The baseline methods either generate poor canonical geometry (w/o LBS Loss) or cannot deform well. In contrast, our method can generate realistic geometry which can also be well deformed to the target poses and expressions.}

\label{fig:ablation}
\end{figure}

How to deform the implicit head avatar is a significant problem in implicit modeling. Deformation under the control of low dimensional meaningful variables is more difficult. Here we illustrate the superiority of the deformation module in GANHead by comparing our method to carefully designed baselines. These baselines are built by replacing our deformation module with the following deformation methods:

\noindent\textbf{Forward skinning for human head (Head-FS).}
Since our deformation module is designed based on the multi-subject forward skinning for human body \cite{chen2022gdna}, we design a baseline that simply applies the forward skinning method to human head. The multi-subject forward skinning method is built upon the human body model SMPL \cite{loper2015smpl}, so we directly change the SMPL model to human head model FLAME \cite{li2017learning} to model the deformation of human head.
As can be observed in the top row of Fig. \ref{fig:ablation}, the model can be well generated, but the generated avatar cannot be deformed to new expressions since the original forward skinning method does not model the non-rigid deformation controlled by expression blendshapes.

\noindent\textbf{FLAME deformation field (F-Def).} F-Def directly uses the pose-dependent corrective bases, expression bases and LBS weights, as well as the standard linear blend skinning of the FLAME head model \cite{li2017learning} to deform the generated avatars. Since FLAME is based on explicit representation, the number of vertices is fixed, we sample the bases and weights to the same number of points as our query points. From Fig. \ref{fig:ablation} (the second row), we observe that the model can generate acceptable canonical shape, but jagged distortion will appear when deforming the avatar.

\noindent\textbf{GANHead deformation module without LBS loss (w/o LBS loss).} The LBS loss plays an important role in the learning of canonical geometry. Here we remove the LBS loss, and the results are shown in Fig.~\ref{fig:ablation} (the third row). It can be observed that the canonical shape is poor, and the geometry is learned in the blendshapes.

\begin{figure}[h]
\centering
\includegraphics[scale=0.342]{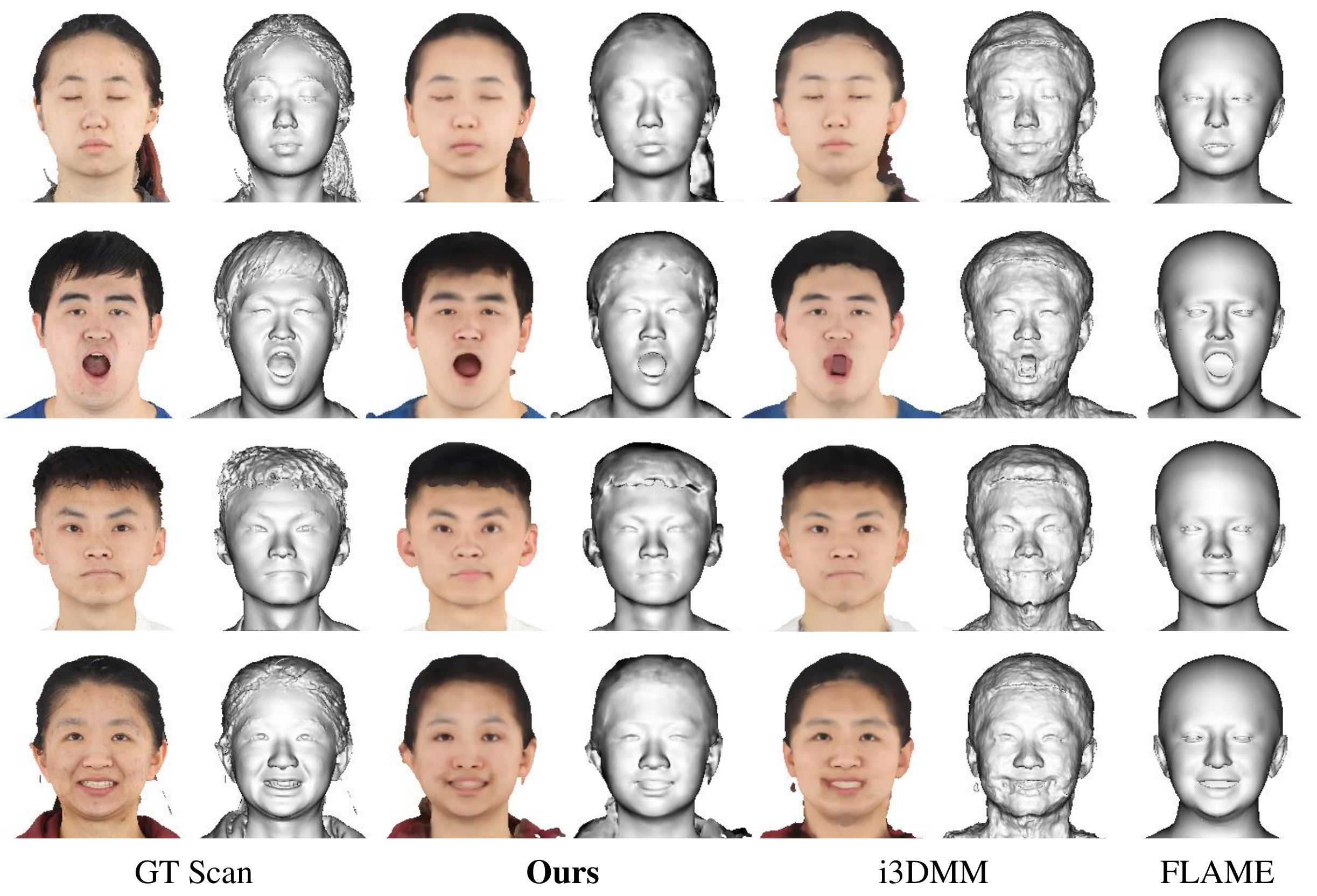}
\vspace{-3mm}
\caption{\textbf{Raw scan fitting.} We compare our fitting results on the test set of FaceVerse-Dataset \cite{wang2022faceverse} with two SOTA head models.}

\vspace{-1mm}
\label{fig:fit}
\end{figure}

\subsection{Comparisons on Scan Fitting}

Although the principal function of GANHead is to generate animatable head avatars with complete geometry and realistic texture, GANHead can also be fitted to raw scans like traditional 3DMMs. In this section, we demonstrate the fitting ability of GANHead through qualitative and quantitative results on the FaceVerse test set \cite{wang2022faceverse}.

Fitting GANHead to raw scans can be achieved by optimizing the shape, detail and color latent codes using the following loss function:
\begin{equation}
\label{eq:fit}
\mathcal{L}_{fit} = \mathcal{L}_{occ} + \mathcal{L}_{3D} + \lambda_{r}\mathcal{L}_{reg},
\end{equation}
where $\mathcal{L}_{occ}$ measures the binary cross entropy between the predicted occupancy and the ground truth occupancy. $\mathcal{L}_{3D}=\lambda_{c} \left \| c_{gt} - c \right \|_2^2 + \lambda_{n}(1 - n_{gt}^{T} \cdot n)$ supervises the reconstructed color and normal of the query points. $\mathcal{L}_{reg}=\lambda_1 \left \| \boldsymbol{z}_{\rm{shape}} \right \|_2^2 + \lambda_2 \left \| \boldsymbol{z}_{\rm{detail}} \right \|_2^2 + \lambda_3 \left \| \boldsymbol{z}_{\rm{color}} \right \|_2^2$ is the regularization term for the three latent codes.

We compare our raw scan fitting results with two SOTA generative head models (i3DMM \cite{yenamandra2021i3dmm} and FLAME~\cite{li2017learning}) that can model complete head and can be animated, which are closest to the objective of this paper. For the fair comparison, we fit FLAME to raw scans by iteratively solving the optimization problem for each scan, and retrain the i3DMM model on the FaceVerse training set.
The qualitative results are shown in Fig.~\ref{fig:fit}. Apparently, our model achieves the best reconstruction quality on both shape (including expression) and texture. FLAME does not model hair, consequently it cannot fit the hair region of raw scans, while i3DMM and our method can model hair region. We further report symmetric Chamfer distance (Ch.) and F-Score for assessing the geometry reconstruction quality, and symmetric color distance for texture assessment, as shown in Tab. \ref{tab:fit}. Our method significantly superior to FLAME and i3DMM in shape and expression reconstruction, especially in the facial region. As for the texture, although i3DMM has slightly better symmetric color distance in the facial region, our method numerically outperforms i3DMM in the full head avatar (head and shoulder) by a margin and achieves a better overall visual effect.

\begin{table}
\begin{center}

\begin{tabu}{c|c|c|c|c}
\tabucline[0.8pt]{-}
Region & Method & Ch. ($\downarrow$)  & F-Score ($\uparrow$) & Color ($\downarrow$) \\
\tabucline[0.8pt]{-}
\multirow{3}{*}{\shortstack{Full \\ Avatar}} & FLAME & 4.883 & 72.78 & -- \\
\cline{2-5}
& i3DMM & 2.583 & 88.49 & 8.819 \\
\cline{2-5}
& \textbf{Ours} & \textbf{2.186} & \textbf{90.37} & \textbf{8.324} \\

\hline \hline

\multirow{3}{*}{Face} & FLAME & 1.755 & 89.59 & -- \\
\cline{2-5}
& i3DMM & 1.208 & 97.30 & \textbf{6.423} \\
\cline{2-5}
& \textbf{Ours} & \textbf{0.695} & \textbf{99.23} & 6.529 \\
\tabucline[0.8pt]{-}

\end{tabu}
\end{center}
\vspace{-3mm}
\caption{\textbf{Fitting comparison.} We report the symmetric Chamfer distance ($\times 10^{-2}$), F-Score computed with a threshold of 0.05, and color distance on the FaceVerse test set.}
\vspace{-3mm}
\label{tab:fit}
\end{table}

\section{Conclusion}
We propose GANHead (\textbf{G}enerative \textbf{A}nimatable \textbf{N}eural \textbf{Head} Avatar model), a novel generative head model that combines the fine-grained control of explicit 3DMMs with the realism of implicit representations. Specifically, GANHead represents coarse geometry, detailed normal and texture via three networks in canonical space to generate complete and realistic head avatars. The generated head avatars can then be directly animated by FLAME parameters via the deformation module. Extensive experiments demonstrate the superiority of GANHead in head avatar generation and raw scan fitting. We further discuss the limitations and broader social impact in Sup. Mat.

\noindent\textbf{Acknowledgments}: This work was supported by NSFC (No. 62225112, 61831015, 62101325, 62201342), the Fundamental Research Funds for the Central Universities, National Key R\&D Program of China (2021YFE0206700), Shanghai Municipal Science and Technology Major Project (2021SHZDZX0102).


{\small
\bibliographystyle{ieee_fullname}
\bibliography{main}
}

\end{document}